# Development of a ROS-based Architecture for Intelligent Autonomous on Demand Last Mile Delivery

Georg Novotny[1,2][0000−0001−8990−2622], Walter Morales-Alvarez[1][0000−0001−6912−4130], Nikita Smirnov[1][0000−0002−6051−9934], and Cristina Olaverri-Monreal[1][0000−0002−5211−3598]

[1] Chair for Sustainable Transport Logistics 4.0, Johannes Kepler University Linz, Austria
{georg.novotny, walter.morales_alvarez, nikita.smirnov_vladimirovich, cristina.olaverri-monreal}@jku.at
www.jku.at/its

[2] UAS Technikum Wien, Höchstaedtplatz 6, 1200 Vienna, Austria
georg.novotny@technikum-wien.at

**Abstract.** This paper presents the development of the JKU-ITS Last Mile Delivery Robot. The proposed approach utilizes a combination of one 3D LIDAR, RGB-D camera, IMU and GPS sensor on top of a mobile robot slope mower. An embedded computer, running ROS1, is utilized to process the sensor data streams to enable 2D and 3D Simultaneous Localization and Mapping, 2D localization and object detection using a convolutional neural network.

**Keywords:** Last Mile Delivery · Mobile Robot · Sensors · Sensor-Fusion · ROS1

## 1 Introduction

The use of mobile robots as delivery aids for postal delivery has seen an upswing in recent years. In addition to Amazon, there are several other manufacturers specializing in "last mile delivery" [16]. The ITS-Chair Sustainable Transport Logistics 4 0. has developed several concepts and solutions to contribute to a more sustainable delivery of goods that requires less traffic [6, 8]. On one hand, this is because the last mile of the delivery accounts for up to 75% of the total supply chain costs [14] and, on the other hand, customer needs and consumer behavior have changed significantly in times of e-commerce and mobile shopping.

Two global megatrends in particular, urbanization and e-commerce, are strong drivers of ever-increasing demand for last-mile delivery services. Urbanization refers to the trend of more and more people moving to urban areas in general and to "megacities" with 10 million inhabitants and more in particular. It is estimated that between 82 and 90% of the world's population, depending on the region, will live in major cities by 2050 [14]. In addition, e-commerce is steadily increasing, and more and more retail goods are being ordered online. In 2021, the revenue of B2C eCommerce in Germany alone grew by 16% and is expected to reach $7.385 billion by 2025 [12]. Thus, greater geographic concentration and increasing online orders per person trigger a rapid increase in the amount of packages that need to be handled. For Germany, for example, it is predicted that 5.68 billion shipments will need to be handled annually by 2025 compared to 2.167 billion in 2012 [4].

The increasing demand for parcels in cities leads to a much higher number of delivery trucks in city centers, which puts additional strain on the existing infrastructure, causes congestion, and negatively impacts on health, environment, and safety. As a result, growing customer awareness and new government regulations force courier services to increase their efforts to operate in a sustainable and environmentally friendly manner. To overcome these challenges, we present an autonomous delivery robot that can navigate in an urban environment. To this end, we developed a software and hardware architecture for a mobile robot for the delivery of packages and letters within the campus of the Johannes Kepler University in Linz Upper Austria (JKU).

The remainder of this paper is structured as follows: In sections 2 and 3 we describe the system concepts, including the hardware, sensor, and software setup. In section 4 we present the results, finally section 5 concludes the paper outlining future research.

## 2   Hardware Setup

Fig. 1 gives an overview of the implemented hardware components which are described in detail in the next section. Although the mobile robot has autonomous capabilities, an operator station is needed to provide a safety fallback and a teleoperation system. In addition, the mobile robot itself needs to be equipped with numerous sensors ranging from 3D LIDAR for obstacle avoidance and mapping, and a front facing camera for obstacle classification as well as for teleoperation.

The LMDBot was built upon a prototype of the "Spider ILD01" slope lawn mower [1] as a base platform which was equipped with a wooden parcel station, to store the packages to be delivered. The original holonomic lawn mower has been transformed into a quasi Ackermann robot in which the chain drive responsible for steering the four wheels has been placed on only two wheels.

### 2.1   Sensor suit

To allow the LMDBot to perceive the environment and move throught it, we provided the robot with the common sensor suit that can be found in autonomous driving. This configuration included several types of sensors whose data guarantee a secure driving through an urban environment (Fig. 1). Specifically the sensors are:

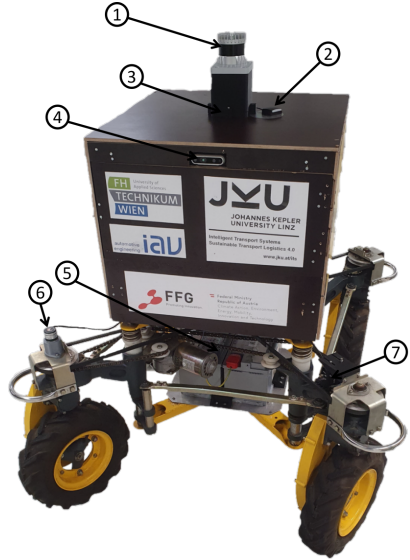

Fig. 1: Last Mile Delivery Robot (LMDBot) hardware setup
(1) Ouster OS1, (2) Ublox C94-M8P, (3) Phidgets Spatial 3/3/3 Inertial Measurement Unit, (4) Intel Realsense D435, (5) 2 × 12V Lead-Acid Batteries, (6) Steering Encoder, (7) Propulsion Encoder

**LIDAR Sensor:** The robot is provided with a Light Detection And Ranging (LIDAR) on the roof that serves to obtain 3D distance information of the environment that is used to localize the robot and detect pedestrians. The selected sensor is the 128 layer LIDAR OS-1 manufactured by Ouster. This sensor has uniform distribution of lights, an effective range of 120 m, a vertical field of view of 45° and a horizontal field of view of 360°. We placed the LIDAR on a Fused Filament Fabrication (FFF) printed platform of 0.3 m over the roof of the robot to minimize the points that are detected due to light ray colliding with the roof of the robot.

**Depth Camera:** We equipped the LMDBot with the depth camera Intel Realsense D435 RGB-D to detect pedestrians and extract dense depth information of the near objects in front of the vehicle. This sensor extracts depth information using two IR cameras for stereo-vision that are overlayed with an IR projector to aid the stereo vision in low light and low feature scenes. Additionally, the sensor possesses a RGB camera that we used to implement an object detection algorithm.

**GNSS INS:** To localize the LMDBot in the environment and track its movements, we provided the robot with the combination of one u-blox GPS [15], one

Phidgets Spatial IMU [10]. With this system we can obtain the position of the robot in local and global coordinates.

**Encoders:** Finally, we also equipped the LMDBot with two FOTEK rotary encoders, one for the propulsion and one for the steering motor. These rotary encoders connect to a low level controller to track the speed and steering of the robot.

### 2.2   Processing

We had to minimize the size and weight of the processing units because the LMDBot main cargo is supposed to be the deliveries that will be placed inside the vehicle. We also had to ensure that the processing units could operate without interruption, given the large volume of data collected by the sensors. For these reasons, we selected two embedded processors, one with a dedicated GPU to ensure quick image data processing and one that requires low energy. The processing units are as follows:

**Main computer** We chose the Nvidia Jetson AGX Xavier Developer Kit [9] to perform the mapping, localization and, path planning. It additionally provided us CUDA capabilites that also allowed us to deploy the deep learning models to perform pedestrian detection.

**Low level control computer** We chose the Raspberry Pi 3B+ for the low level control due to its simplicity. It communicated with the main computer via ethernet and received the speed and steering commands from the main computer. The low level controller used these commands to calculate the amount of voltage that was needed by the motor actuators of the robot to achieve the desired speed and steering commands.

### 2.3   Network Communication

To transfer data between the LIDAR the main computer and the low level control computer we used an router of 1 Gbps per channel. The router allows the different components of the system to communicate using the TCP/IP protocol. On the other hand, the GPS INSS and the camera interface with the main computer through USB 3.2 which provides rapid data transmission.

### 2.4   Power management

We equipped the robot with two batteries of 95 AH and 850 A peak current each, since the actuators of the robot require 24V DC to operate, and require high current due to the the robot's weight. To segregate the power channels, we linked the robot's components via a fuse box for safety. We also attached a switch to turn the robot on and off, as well as a secondary switch to the DC motors to stop the robot without turning it off. Finally, a circuit breaker safeguards the systems from current spikes that might damage the hardware. The connection diagram can be seen in Fig. 2.

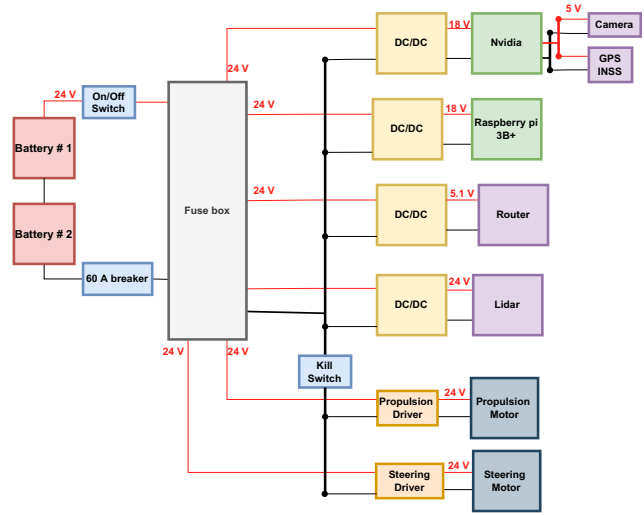

Fig. 2: Power management diagram of the robot.

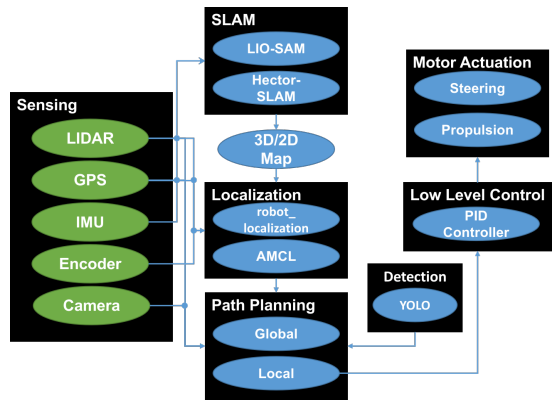

Fig. 3: Overview of software components and interaction between them

## 3  Software Architecture

The following section will provide information about the various software components utilized on the LMDBot.

### 3.1 ROS Architecture

The Robot Operating System (ROS) [11] is used as high-level API to evaluate sensor data and control actuators via our developed low-level control either via keyboard or joystick inputs. The Fig. 3 visualizes an overview of the implemented software components.

**Sensor Calibration:** The Intel Realsense D435 RGB-D is a camera whose intrinsic parameters are already provided by the manufacturer. As a result, there was no need to use any intrinsic calibrator package. The extrinsic calibration, providing translation in x, y, and z as well as roll, pitch and yaw between the camera and the LIDAR, was created following the algorithm described in [1]. To ease the extrinsic calibration of the IMU we mounted it exactly below the origin of the LIDAR using the aforementioned FFF printed structure.

**Detection:** To detect objects that lie along the path of the robot, we applied the Convolutional Neural Network (CNN) *YOLOV4* [2] to the image stream from the RGB-D camera. To improve the adaptability of the CNN to our use case, we re-configured it to detect only people, dogs, cats, ducks, scooters, and bicyclists, as these are the main dynamic objects on the campus of the JKU.

**Mapping:** The mapping process of the campus was performed using two different methods. On the one hand, we performed classical 2D mapping with *Hector-SLAM* [5], on the other hand, we created a 3D map using *LIO-SAM* [13]. To create the mandatory 2D LIDAR for *Hector-SLAM* we cut the 3D information of the Ouster OS1 to a 2D plane utilizing the *pointcloud_to_laserscan* ROS package.

**Localization:** The localization was done based on the *Adaptive Monte Carlo localization (AMCL)* [3] as well as the *robot_localization* package. *AMCL* takes over the global localization in the 2D map and the *robot_localization* package fuses the sensor data of the wheel encoders, IMU and the GPS signal by means of Extended Kalman Filter (EKF) and then feeds them into *AMCL*.

**Low Level Control:** To control our robot, we implemented a PID controller for the drive motor and a PID controller for the control motor, where the controlled distance was the linear and angular velocity in x and around the z axis, respectively. For initial tuning of the PID parameters we relied on Visual Odometry from RTAB-Map [7]. Furthermore, for low-level control, we simplified our vehicle model to that of a bicycle with the origin lying in the middle of the rear axis.

## 4 Results

A 2D map was generated by relying on the Hector-SLAM [5]. The corresponding 3D map was produced using LIO-SAM [13]. Finally, the objects in the vicinity were detected relying on the YOLO CNN V4 [2]. The results are visualized in Fig. 4.

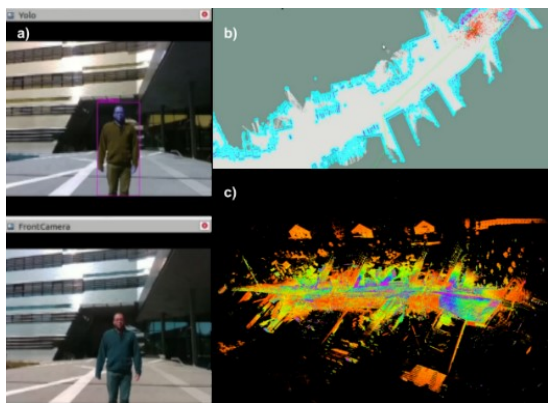

Fig. 4: a) Object detection, b) Generated 2D map of the JKU campus, c) Dense 3D map of the JKU campus

As visible in Fig. 4 b) the 2D map is quite noisy, we believe the high number of glass fronts in combination with dynamic objects (pedestrians) played a significant role here. As figure 4 c) depicts the generated 3D map extends far beyond the actual campus of JKU, which can result in better localization as the buildings in the background can be used as landmarks.

## 5 Conclusion and Outlook

In this paper a prototype of a last mile delivery robot has been presented. We introduced our hardware as well as software stack and presented results in terms of generated 3D and 2D maps.

Further work will deal with creating more precise maps, to ease the path planning, and evaluating the performance of autonomous delivery between two or more positions in the geographic coordinate system available at the JKU campus. Further, we will investigate route optimization methods for the parcel delivery framework.

**Acknowledgements** This work was supported by the Austrian Ministry for Climate Action, Environment, Energy, Mobility, Innovation and Technology (BMK) Endowed Professorship for Sustainable Transport Logistics 4.0., IAV France S.A.S.U., IAV GmbH, Austrian Post AG and the UAS Technikum Wien.